\documentclass[letterpaper]{article} 
\usepackage{aaai24}  
\usepackage{times}  
\usepackage{helvet}  
\usepackage{courier}  
\usepackage[hyphens]{url}  
\usepackage{graphicx} 
\usepackage{natbib}  
\usepackage{caption} 
\usepackage{algorithm}
\usepackage{algorithmic}

\usepackage{epsfig}
\usepackage{graphicx}
\usepackage{amsmath}
\usepackage{amssymb}
\usepackage{xcolor}
\usepackage{colortbl}
\usepackage{array}
\usepackage{enumitem}
\usepackage{multirow}
\usepackage{booktabs}
\usepackage{amsmath,bm}
\usepackage{pifont}
\usepackage{colortbl}
\usepackage{array}

\usepackage{newfloat}
\usepackage{listings}
\DeclareCaptionStyle{ruled}{labelfont=normalfont,labelsep=colon,strut=off} 
\floatstyle{ruled}
\newfloat{listing}{tb}{lst}{}
\floatname{listing}{Listing}
\title{Cycle-Consistency Learning for Captioning and Grounding}
\author{
	Ning Wang\textsuperscript{\rm 1},~
	Jiajun Deng\textsuperscript{\rm 2},~
	Mingbo Jia\textsuperscript{\rm 1}\\
}
\affiliations{
	\textsuperscript{\rm 1}Huawei Inc.~~~\textsuperscript{\rm 2}University of Adelaide, Australian Institute for Machine Learning\\
	wn6149@mail.ustc.edu.cn,~
	jiajun.deng@adelaide.edu.au,~
	jiamingbo@huawei.com 
}

\usepackage{bibentry}

\begin{document}

\maketitle

\begin{abstract}
We present that visual grounding and image captioning, which perform as two mutually inverse processes, can be bridged together for collaborative training by careful designs.
By consolidating this idea, we introduce CyCo, a cyclic-consistent learning framework to ameliorate the independent training pipelines of visual grounding and image captioning.
%
The proposed framework (1) allows the semi-weakly supervised training of visual grounding; (2) improves the performance of fully supervised visual grounding; (3) yields a general captioning model that can describe arbitrary image regions.
Extensive experiments show that our fully supervised grounding model achieves state-of-the-art performance, and the semi-weakly supervised one also exhibits competitive performance compared to the fully supervised counterparts.
Our image captioning model has the capability to freely describe image regions and meanwhile shows impressive performance on prevalent captioning benchmarks.
\end{abstract}

\section{Introduction}

The recent decades have witnessed the great success in vision-language (VL) related fields.
Based on the primary target of bridging the modality gap between vision and language, deep neural networks addressing VL tasks generally share the pre-training objective, model structure, large-scale training corpus, \emph{etc}. 
However, by the time of downstream fine-tuning, these tasks are typically individually tackled or simply combined in a multi-task training paradigm.

\begin{figure}
	\centering
	\includegraphics[width=7.9cm]{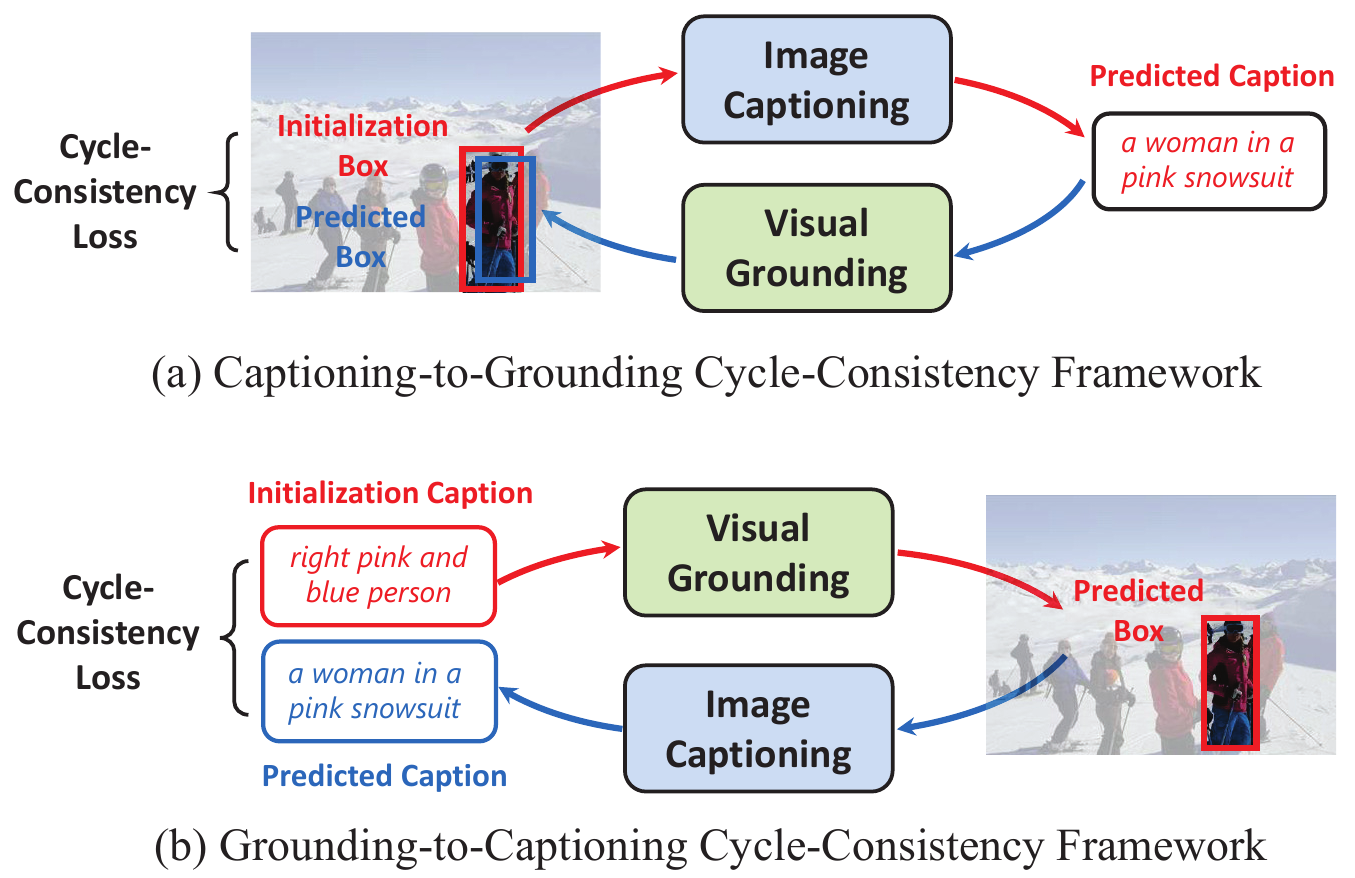}
	\caption{The proposed framework jointly optimizes image captioning and visual grounding models via cycle-consistency learning. In (a), our method computes the region (box) consistency in the captioning-to-grounding optimization cycle. In (b), our framework measures the caption consistency in the grounding-to-captioning cyclic process.}
	\label{fig:intro}
\end{figure}

In this work, we devote our efforts to two VL downstream tasks including image captioning \cite{show-and-tell,BUTD} and visual grounding \cite{RefCOCOg,RefCOCO/+}, and explore their inherent relationships to enable effective joint training. To match the granularity of visual parts in visual grounding, we first extend image captioning to a more general scenario, where the model is intended to describe a given region.
%
%
%
We define this generalized task as regional image captioning, which is similar to dense captioning task \cite{johnson2016densecap} but is free of the requirement of object detection.
Particularly, the conventional task definition of image captioning \cite{show-and-tell} is a special case of regional image captioning that regards the whole image as a region.
%
For visual grounding,
which is also known as referring expression comprehension \cite{RefCOCOg,RefCOCO/+} and phrase localization \cite{Referitgame,Flickr30k} in the literature, we maintain the original task target to localize the corresponding region (generally denoted by a bounding box) described by a given language expression.

Despite achieving inspiring progress, image captioning and visual grounding still suffer from several limitations.
For visual grounding, this task simultaneously requires text descriptions and accurate object bounding boxes for model optimization.
These fine-grained image-text-box triplets are rather laborious to collect.
How to optimize the grounding model using limited data annotations has received considerable attention \cite{Relation-aware-weak-grounding,contrastive-weak-grounding}.
As for image captioning, existing methods typically focus on describing the whole image.
%
%
The capabilities of modeling region relationships and properly describing them are largely overlooked in existing algorithms.
We argue that a robust image captioner should be qualified to freely describe the image, from an arbitrary region to the whole image.
Similar to the grounding task, obtaining the regional captioning ability also requires sufficient image-text-box data, increasing the training cost.

%
In this paper, we introduce a joint learning framework, namely CyCo, to ameliorate the training pipelines of image captioning and visual grounding via cyclic consistency.
%
%
Our core motivation is that visual grounding and regional image captioning can be regarded as an inverse process of each other.
Specifically, a visual grounding model takes an image and a region-specific description as inputs to predict the position of the corresponding bounding box, while regional image captioning receives the location of a region to produce a region-aware caption.
%
%
When taking one's output as the input of the other, these two tasks naturally establish a cyclic structure.
%
%
%
As shown in Figure~\ref{fig:intro} (a), the generated bounding box from the grounding model is expected to be consistent with the input box of the regional image captioner.
In this process, the whole framework merely needs an initialized bounding box for training.
As depicted in Figure~\ref{fig:intro} (b), the produced text description from the regional image captioner is expected to be consistent with the input referring text of the grounding model.
As a result, the models in this cycle are free of the bounding box annotations.

The proposed cycle-consistency learning framework enjoys the following merits.
(1) We bridge two independent VL tasks in a unified framework. To this end, our method can potentially absorb the training data from both tasks and share the model parameters for collaborative training.
(2) Thanks to the joint training and online data augmentation (iterative pseudo label generation) in the cyclic learning process, under the \emph{same} training data, our framework further improves the performance of the supervised grounding model.
(3) After collaborative training, our framework yields a strong image captioning model that can describe the visual contents in different spatial levels, \emph{i.e.}, from a subregion to the global image.
(4) The proposed framework allows the semi-weakly supervised training of the grounding model, which merely needs limited fully-annotated images and many more images with only bounding box annotations \emph{or} only language expression labels for model training.

In summary, we make three-fold contributions:
\begin{itemize}[noitemsep,nolistsep]	
	\item[$\bullet$] We present a novel cycle-consistency learning framework to bridge two independent vision-language tasks including image captioning and visual grounding.
	\item[$\bullet$] We design a simple regional image captioner and a Transformer-based grounding model. We further organize them in a unified framework with weight-sharing architecture for efficient end-to-end learning.
	\item[$\bullet$] Extensive experiments validate the effectiveness of our proposed cycle-consistency learning framework.
\end{itemize}

\section{Related Work}

{\noindent \bf Vision-language Pre-training.}
Vision-language (VL) pre-training algorithms \cite{CLIP,OSCAR,BLIP} aim to bridge the domain gap between vision and language representations.
The recent dual-encoder methods such as CLIP \cite{CLIP} align the representations using contrastive learning.
Despite the outstanding performance, their light interaction manner fails to deeply fuse VL representations for generation tasks.
In contrast, recent VL pre-training approaches \cite{VLP,OSCAR,UNIMO,BLIP} adopt a relatively heavy Transformer architecture \cite{Transformer} to achieve the deeper multi-modal interaction.
%
%
%
Inspired by the success of previous arts, we also conduct the cross-modal pre-training to prompt the downstream VL tasks.

{\noindent \bf Visual Grounding.} Traditional visual grounding methods typically follow a two-stage pipeline, which generates plentiful region proposals in the first stage and selects the most matched one via language expression in the second stage \cite{DGA,NMTree}.
Recently, one-stage visual grounding approaches gain increasing attention.
%
They generally embed the linguistic information into the one-stage object detector \cite{FAOA} or model multi-modal representations via Transformer \cite{TransVG,deng2023transvg++} for efficient visual grounding.

\begin{figure*}
	\centering
	\includegraphics[width=17.0cm]{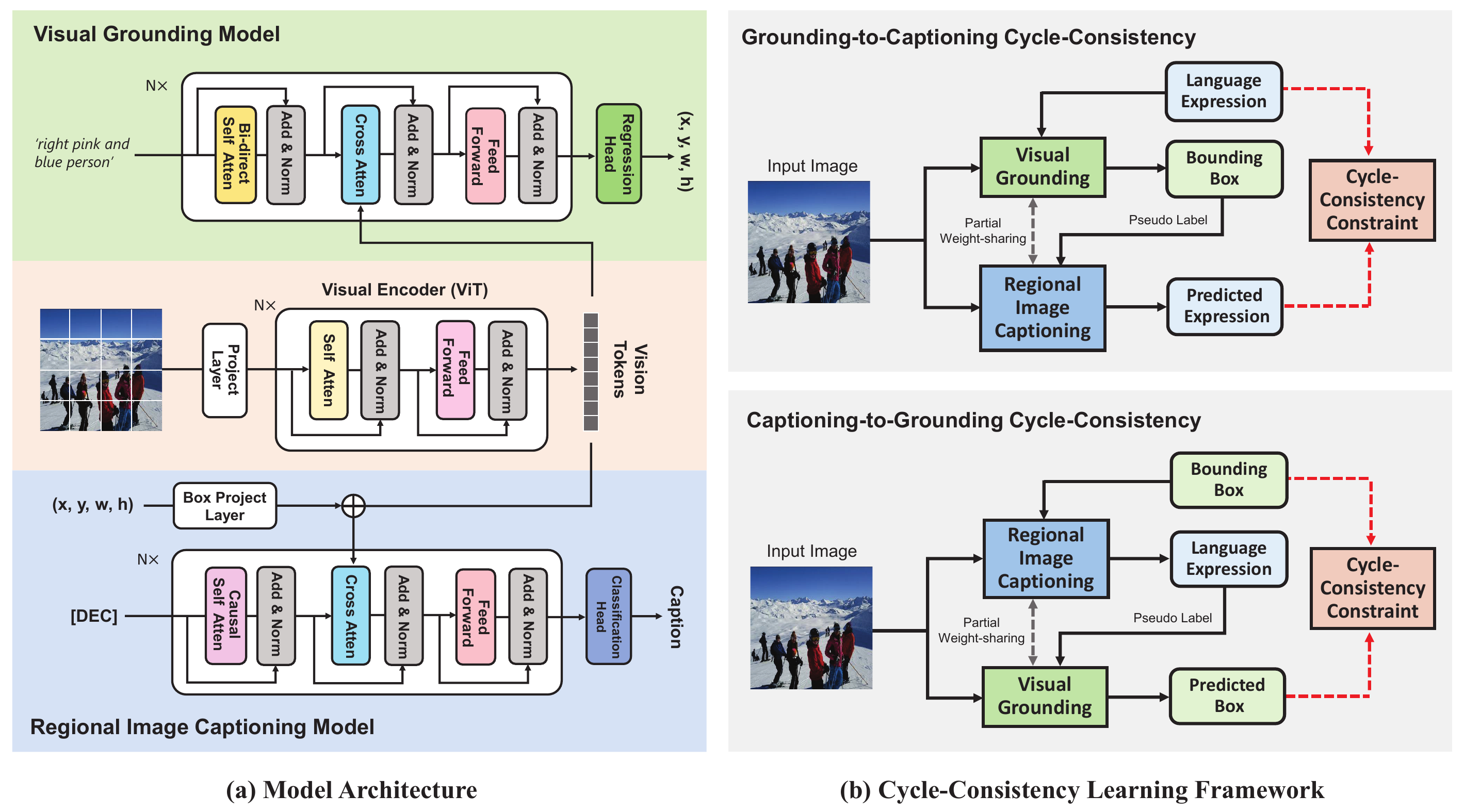}
	\caption{In (a), we exhibit the model architecture of our proposed joint training framework of visual grounding and regional image captioning. Two models share a visual encoder (ViT) and leverage different Transformer blocks for individual tasks. In (b), we show the cyclic consistency learning processes of our framework including a grounding-to-captioning cycle and a captioning-to-grounding cycle.}
	\label{fig:main}
\end{figure*}

Different from the above algorithms based on supervised training, weakly supervised grounding models learn the region-phrase correspondence with only language expressions.
These methods first obtain a set of ROIs (Region of Interest) using object detectors, and then mine the correspondence between ROIs and query expressions for model training \cite{contrastive-weak-grounding,Relation-aware-weak-grounding}. 
In this work, we train the grounding model in a semi-weakly supervised manner with the help of a captioning model.
Our framework is free of external object detectors for data pre-processing.

{\noindent \bf Image Captioning.} 
Image captioning aims to generate a human-readable sentence to describe the image contents.
Captioning algorithms \cite{AoANet,BUTD,VIVO} typically utilize the object detectors to extract ROI features or simply exploit the grid features for efficient visual representation modeling \cite{SimVLM,BLIP}.
After visual feature extraction, captioning models utilize a decoder such as Transformer to generate the sentence \cite{PureT_AAAI,wang2023efficient,wang2023controllable}.
Previous algorithms exploit the region information to facilitate the image captioning \cite{say_as_you_wish,region-caption,show_control_tell}, but they still focus on describing the \emph{global} image content.
The dense captioner \cite{johnson2016densecap} mainly focuses on detecting and describing \emph{local} ROI regions.
In contrast, our image captioner is designed to freely describe \emph{both} global and regional contexts.

Recently, some methods \cite{OFA,yu2017joint} jointly train the captioning and grounding models in a multi-task fashion. 
Mao \emph{et al.} \cite{RefCOCOg} utilize the grounding model as a verification module to push the generated referring expressions to be unambiguous.
%
%
%
%
Different from previous arts, our method bridges the captioning and grounding models in a \emph{cyclic framework} and explores \emph{two} different cycle-consistency constraints for collaborative training.

{\noindent \bf Cycle-Consistency Learning.}
To bridge one modality or domain to the other, cycle-consistency learning has been explored extensively in visual object tracking \cite{UDT,CycleTime}, machine translation \cite{dual_translation}, unpaired image-to-image translation \cite{CycleGAN}, visual question answering \cite{cycleVQA}, image captioning \cite{mscap}, \emph{etc}.
%
%
Different from the previous arts that explore the consistency within a single modality or a single task, we jointly optimize two VL tasks to form different cycles.
To our knowledge, the cyclic consistency between image captioning and visual grounding has rarely been touched in the literature.
Further, we explore the potential of our framework in different training scenarios including semi-weakly supervised and fully supervised training.

\section{Methodology}
Our method follows the pretrain-and-finetune paradigm.
At the pre-training stage, we leverage the widely-adopted training objectives (\emph{i.e.,} image-text contrastive loss, matching loss, and language modeling loss) to align and fuse the visual and linguistic representations.
At the fine-tuning stage, both visual grounding and image captioning models reuse the pre-trained model architecture, while capitalizing on task-specific head networks.
%
During fine-tuning, we further develop the cycle-consistency learning framework to jointly optimize the grounding and captioning models. 
The detailed model architecture and our proposed cycle-consistency learning framework are illustrated in Figure~\ref{fig:main}.

%
%
%

\subsection{Revisiting Model Pre-training}\label{method: pretraining}
%
%
Our pre-trained vision-language model follows the BLIP approach \cite{BLIP}. 
We briefly review its vision encoder, image-grounded text encoder, and image-grounded text decoder, which are highly related to our downstream tasks.

{\noindent \bf Vision Encoder.} We exploit the commonly used ViT-B/16 network \cite{ViT} as the vision encoder. ViT-B/16 is composed of a stack of 12 Transformer encoder layers with 12 heads in each multi-head attention layer.
%
%
Given an image, we first split it into small patches with equally $16\times16$ size, and project them into feature vectors, which are known as vision tokens. We denote the final encoded vision tokens as ${\boldsymbol v}$.
%

{\noindent \bf Image-Grounded Text Encoder.} 
%
%
In this block, we first append a special \texttt{[ENC]} token at the beginning of the text sequence.
Then the text tokens are converted to embeddings via a word embedding layer.
This text encoder leverages the bi-directional self-attention to further encode the text embeddings, and aggregate the visual information through cross-attention with vision tokens for multi-modal fusion.

The output embedding of \texttt{[ENC]} contains the rich multi-modal representation of the image-text pair, which is exploited to compute the image-text matching (ITM) loss.
Specifically, we add a binary classification head on top of it to predict whether an image-text pair is matched.

{\noindent \bf Image-Grounded Text Decoder.} 
This text decoder block is similar to the above image-grounded text encoder, while replacing the bi-directional self-attention with the causal self-attention to facilitate the text generation.
A special \texttt{[DEC]} token is used as the beginning signal of a sequence.

The image-grounded text decoder is optimized by the language modeling (LM) loss, which maximizes the likelihood of the text in an autoregressive manner.
%

\subsection{Cycle-Consistency Learning Framework} \label{method: cycle-consistency}

In this section, we first introduce how to transfer the pre-trained model to downstream visual grounding and regional image captioning tasks. Then, we exhibit how to jointly optimize them by virtue of the cyclic constraints.

{\noindent \bf Visual Grounding (\texttt{VG}).} 
%
%
To facilitate the following introduction of cycle-consistency learning, we depict the visual grounding model in a holistic view, which takes the visual features $\boldsymbol{v}$ and description tokens $\boldsymbol{x}$ as its inputs, and outputs a predicted bounding box ${\boldsymbol{b}_{pred}}$ as follows.
\begin{equation}\label{eq: vg model}
	{\boldsymbol{b}_{pred}} = \texttt{Model}_{\texttt{VG}}(\boldsymbol{v}, \boldsymbol{x}).
\end{equation}
In our framework, visual grounding model $ \texttt{Model}_{\texttt{VG}} (\cdot,\cdot)$ reuses the image-grounded text encoder block, which fuses the visual and text features to predict the region localization.
%
%
Following the setting in the pre-training stage, we also add a special \texttt{[ENC]} token at the beginning of the tokenized tokens.
Then, the bi-directional self-attention encodes the text tokens and cross-attention injects visual information into the text tokens. 
%
%
After multi-layer visual-linguistic fusion via the Transformer structure, the output embedding of the \texttt{[ENC]} token contains rich cross-modal contexts, which is leveraged for bounding box regression via a regression head.
The regression head is a simple three-layer multi-layer perceptron (MLP) with ReLU activations between layers. The output of the box regression head is the 4-dim box coordinates $\boldsymbol{b}_{pred} = (x,y,w,h)$.

In the model fine-tuning stage, we normalize the ground-truth region bounding box by the scale of the image to obtain $\boldsymbol{b}_{gt} = (\hat{x},\hat{y},\hat{w},\hat{h})$, and combine the generalized IoU (GIoU) loss ${\cal L}_{\text{GIoU}}(\boldsymbol{b}_{pred},\boldsymbol{b}_{gt})$ \cite{GIoU} and smooth L1 loss ${\cal L}_{\text{smooth-L1}} (\boldsymbol{b}_{pred},\boldsymbol{b}_{gt})$ to optimize the grounding model.

{\noindent \bf Regional Image Captioning (\texttt{IC}).} 
Similar to the visual grounding, we also formulate the regional image captioning model $\texttt{Model}_{\texttt{IC}} (\cdot,\cdot)$ as a black box, which takes image features $\boldsymbol{v}$ and a specific region (denoted by the box coordinate $\boldsymbol{b}$) as inputs to predict the region-related language expression ${\boldsymbol{x}_{pred}}$ as follows.
\begin{equation}\label{eq: ic model}
	{\boldsymbol{x}_{pred}} = \texttt{Model}_{\texttt{IC}} (\boldsymbol{v}, \boldsymbol{b}).
\end{equation}

This regional image captioner mainly reuses the image-grounded text decoder in the pre-training stage to generate the language expression.
After model pre-training with language modeling (LM) loss, the text decoder already has the zero-shot captioning capability to some extent.
%
%
Nevertheless, different from the classic image captioner, our model is required to be region-aware.
To this end, in the fine-tuning stage, we project the box coordinate $\boldsymbol{b}$ to the regional embedding via a fully-connected layer: $\boldsymbol{e}_{box} = \texttt{FC}_{\texttt{box}}(\boldsymbol{b})$.
%
%
This regional embedding is added to the vision tokens to obtain the region-aware visual representations $\boldsymbol{v}^{\star}$:
\begin{equation}\label{key}
	\boldsymbol{v}^{\star} = \boldsymbol{v} + \boldsymbol{e}_{box}.
\end{equation}

%
Different from the bi-directional self-attention in visual grounding, the captioning model utilizes causal self-attention to facilitate text generation. As shown in Figure~\ref{fig:main} (a), a classification head upon Transformer is used to generate tokens over the vocabulary.
For a text sequence $\boldsymbol{x} = \{x_1, x_2, \cdots, x_n\}$, image captioner generates token $x_t$ based on the previous tokens ${\boldsymbol x}_{<t}$ in an auto-regressive fashion.
We train the captioning model using unidirectional language modeling objective by maximizing the negative log-likelihood of the text sequence: $- \sum_{t} \text{log}P_{\theta} \left( x_t| \boldsymbol{v}^{\star}, {\boldsymbol x}_{<t} \right)$,
where $\theta$ denotes the trainable model parameters.

{\noindent \bf \texttt{VG} $\rightarrow$ \texttt{IC} Cycle-Consistency Learning.}
As shown in Eq.~\ref{eq: vg model} and Eq.~\ref{eq: ic model}, visual grounding and regional image captioning perform as the inverse process of each other.
Consequently, we can organize them in a cyclic framework for joint optimization using consistency constraints.
In the fine-tuning stage, inspired by BLIP \cite{BLIP}, grounding and captioning branches share the model parameters of cross-attention and feed-forward network (FFN) in their Transformer architectures to tightly bridge two individual tasks.
This weight-sharing mechanism not only improves training efficiency but also enjoys multi-task learning for mutual prompting.

We first start the cycle-consistency learning from visual grounding (\texttt{VG}) to image captioning (\texttt{IC}), and leverage the training objective of image captioning to optimize two tasks.
This process merely needs the language expression $\boldsymbol{x}_{init}$ and visual features $\boldsymbol{v}$ as the inputs, without requiring any bounding box labels as follows.
\begin{align}
{\bf \texttt{VG:}~~~}& {\boldsymbol{b}_{pred}} = \texttt{Model}_{\texttt{VG}}(\boldsymbol{v}, \boldsymbol{x}_{init}), \\
{\bf \texttt{IC:}~~~}& {\boldsymbol{x}_{pred}} = \texttt{Model}_{\texttt{IC}}(\boldsymbol{v}, \boldsymbol{b}_{pred}), \\
{\bf \texttt{Loss:}~~~}& {\cal L}_{\texttt{VG} \rightarrow \texttt{IC} } = {\cal L}_{\text{XE}}(\tilde{\boldsymbol{x}}_{init}, \tilde{\boldsymbol{x}}_{pred}),
\end{align}
where ${\cal L}_{\text{XE}}(\cdot,\cdot)$ denotes the cross-entropy loss, $\tilde{\boldsymbol{x}}_{init}$ represents the one-hot vocabulary distribution of the input expression ${\boldsymbol{x}}_{init}$, ${\boldsymbol{x}}_{pred}$ denotes the predicted text sequence and $\tilde{\boldsymbol{x}}_{pred} $ is the corresponding token prediction probability.
In this cycle-consistency learning process, we first utilize the grounding model $\texttt{Model}_{\texttt{VG}}(\cdot,\cdot)$ to generate the bounding box coordinate ${\boldsymbol{b}_{pred}}$, which serves as the pseudo label of captioning model $\texttt{Model}_{\texttt{IC}}(\cdot,\cdot)$.
Then we can utilize the initial language expression ${\boldsymbol{x}}_{init}$ as the supervision signal of the generated caption ${\boldsymbol{x}}_{pred}$ to form the cyclic supervision constraint.
%
In this way, we can optimize the model without providing any bounding box annotations.

{\noindent \bf \texttt{IC} $\rightarrow$ \texttt{VG} Cycle-Consistency Learning.} We can also build the cycle-consistency learning from image captioning (\texttt{IC}) to visual grounding (\texttt{VG}). 
This cycle merely requires the visual feature and an initial bounding box $\boldsymbol{b}_{init}$ as the inputs:
\begin{align}
{\bf \texttt{IC:}~~~}&  {\boldsymbol{x}_{pred}} = \texttt{Model}_{\texttt{IC}}(\boldsymbol{v}, \boldsymbol{b}_{init}), \\
{\bf \texttt{VG:}~~~}&  {\boldsymbol{b}_{pred}} = \texttt{Model}_{\texttt{VG}}(\boldsymbol{v}, \boldsymbol{x}_{pred}), \\
{\bf \texttt{Loss:}~~~}&  {\cal L}_{\texttt{IC} \rightarrow \texttt{VG} } = {\cal L}_{\text{GIoU+L1}}(\boldsymbol{b}_{init}, \boldsymbol{b}_{pred}),
\end{align}
where ${\cal L}_{\text{GIoU+L1}} (\cdot,\cdot)$ denotes the combination of generalized IoU loss \cite{GIoU} and smooth L1 loss.
Ideally, the generated bounding box ${\boldsymbol{b}_{pred}}$ should be consistent with the initial bounding box $\boldsymbol{b}_{init}$. To this end, we can optimize the whole network via the visual grounding loss.

\subsection{Model Training} \label{method: training details}
Both visual grounding and regional image captioning models require the image-text-box triplets for training. 
Labeling such well-annotated data is rather time-consuming.
However, the aforementioned two learning cycles enable us to optimize the models in a semi-weakly supervised manner, where either the ground-truth language expression or bounding box can be omitted.
Besides, based on the same data, we observe that adding cycle-consistency constraints to the classic fully supervised training paradigm can further boost the performance, which shows our cycle-consistency learning framework can better exploit the training data.

{\noindent \bf Fully Supervised Training.}
In the supervised training, except for the individual training objectives for grounding and captioning models, we also add the cycle-consistency losses (\emph{i.e.,} ${\cal L}_{\texttt{VG} \rightarrow \texttt{IC}}$ and ${\cal L}_{\texttt{IC} \rightarrow \texttt{VG}}$) to regularize the models.

The effectiveness of our framework can be explained in an online data augmentation view.
It is well recognized that a picture is worth thousands of words. 
The proposed cycle-consistency learning framework can be regarded as an incremental training process that iteratively predicts pseudo labels to augment the training data (\emph{e.g.,} diverse referring expressions for a region), and thus further boosts the performance of supervised training.

{\noindent \bf Semi-weakly Supervised Training.}
We also validate the potential of our framework by conducting semi-weakly supervised training, where only limited fully-annotated images (with \emph{both} referring expression and bounding box) are available.
In the experiments, we empirically set the percentage of fully-annotated data as 20\%, and the rest 80\% images are annotated with only language expressions \emph{or} only bounding boxes.
For fully-annotated images, we compute the standard fully-supervised losses for the 20\% image-text-box data.
As for the partially labeled data, we compute the \texttt{IC} $\rightarrow$ \texttt{VG} consistency loss for box-only images, and compute the \texttt{VG} $\rightarrow$ \texttt{IC} consistency loss for text-only images. All the losses are gathered as the total loss of a batch of training samples.
%
%
By virtue of \texttt{VG} $\rightarrow$ \texttt{IC} and \texttt{IC} $\rightarrow$ \texttt{VG}, the grounding and captioning models mutually annotate the weakly-annotated data for collaborative training.
Note that the captioning and grounding models share the same visual encoder (ViT) and most blocks (cross-attention and FFN) of their Transformer blocks.
To this end, they both benefit from the \texttt{VG} $\rightarrow$ \texttt{IC} and \texttt{IC} $\rightarrow$ \texttt{VG} cycle-consistency training.


%
%
%
%

\section{Experiments}

\subsection{Datasets and Metrics}

{\noindent \bf Pre-training Data.}
In the pre-training stage, we collect the image-text pairs from Visual Genome \cite{VG}, COCO \cite{MSCOCO}, SBU \cite{SBU}, Conceptual 3M \cite{CC3M}, and a filtered version of LAION (115M images) \cite{LAION}.

{\noindent \bf RefCOCO, RefCOCO+, and RefCOCOg.} 
We evaluate the visual grounding performance on three prevalent benchmarks including RefCOCO \cite{RefCOCO/+}, RefCOCO+ \cite{RefCOCO/+}, and RefCOCOg~\cite{RefCOCOg}.
%
%
%
Following the official setting, RefCOCO and RefCOCO+ are split into the train set, validation set, testA set, and testB set.
%
%
RefCOCOg includes the train set, validation set, and test set.

We consider a referring expression grounded correctly when its predicted region has at least 0.5 Intersection-over-Union (IoU) with the ground-truth box.
We measure the visual grounding performance in terms of top-1 accuracy.

{\noindent \bf COCO Caption.} We evaluate the image captioning performance on the COCO Karpathy split dataset \cite{MSCOCO,karpathy}.
%
%
To evaluate the performance, we leverage standard metrics in the captioning task including BLEU@4 \cite{BLEU}, METEOR \cite{METEOR}, CIDEr \cite{CIDER}, and SPICE \cite{SPICE}.

\subsection{Implementation Details}
In our framework, the image encoder is initialized from ViT-B/16 pre-trained on the ImageNet \cite{ViT}, and the text encoders of both visual grounding and image captioning branches are initialized from the official BERT-base \cite{BERT}. 
In the pre-training stage, the model is trained on 32 V100 GPUs for 20 epochs using a batch size of 2880. 
We use AdamW optimizer \cite{Adam} with a weight decay of 0.05. 
The learning rate is warmed-up to $3 \times 10^{-4} $ and decayed linearly with a rate of 0.85. 
We take random image crops of resolution $224 \times 224$ during pre-training.

In the fine-tuning stage, we train the model using a small learning rate of $ 1 \times 10^{-5} $ and linearly decay it.
The additionally added blocks that are not included in the pre-training stage (\emph{i.e.,} box regression head and box project layer) are randomly initialized. 
%
For fair comparisons, following \cite{TransVG,BLIP}, the input image resolutions are set to $640 \times 640$ and $384 \times 384$ when evaluating grounding and captioning tasks, respectively.
%
When combining different datasets, we carefully check the train/test sets to avoid image overlap.
%
%
The captioning model adopts the beam search strategy (beam size = 3) in all experiments.
The proposed cycle-consistency model is fine-tuned for 20 epochs.

In the following experiments, our \underline{Cy}cle-\underline{Co}nsistency learning of captioning and grounding framework is denoted as {\bf CyCo}.
%

\subsection{Ablation Study}

{\noindent \bf Semi-weakly Supervised Training.} 
Thanks to the cycle-consistency design, our approach is able to train the grounding model using weakly labeled data.
As shown in Table~\ref{table:ablative on semi-weakly} (top), the grounding performance is poor when only 20\% training data is available.
The performance gap between the models trained using 20\% data and 100\% data is considerable, \emph{e.g.,} 13.4\% gap on RefCOCOg test set. 
%
%
In ``pseudo label'' setting in Table~\ref{table:ablative on semi-weakly}, we train a grounding/captioning model using the 20\% fully-annotated data and leverage this model to label the rest 80\% images. 
Then, all the 100\% data are used to train the model, whose performance can be improved but is still unsatisfactory.
%
%
Further, by annotating the rest 80\% training data in an online manner using our CyCo framework, the performance gap is significantly reduced.
Finally, our semi-weakly supervised CyCo model only has a minor gap in comparison to the fully-supervised model.

In Table~\ref{table:ablative on semi-weakly} (bottom), we further ablative the regional image captioning task, which requires the model to describe a specific region. 
Using only 20\% full-annotated data, our semi-weakly supervised captioning model significantly reduces the performance gap in comparison to the supervised counterparts.
Our CyCo framework has the potential of absorbing more weakly-labeled data such as object detection images with only box labels or image-text pairs with only caption annotations to further improve the performance.

\setlength{\tabcolsep}{2pt}
\begin{table}
\small
\begin{center}
	\caption{Comparison results of semi-weakly supervised and fully supervised models. {Top}: comparison of visual grounding performance in terms of top-1 accuracy. {Bottom}: comparison of regional image captioning performance in METEOR (M) and CIDEr (C).} \label{table:ablative on semi-weakly}	
	
	\begin{tabular*}{8.4 cm} {@{\extracolsep{\fill}}cc|ccc|cc}
		\hline
		\multicolumn{7}{c}{Ablation Study of Visual Grounding} \\
		\hline
		\multicolumn{2}{c|}{Training Data}  & \multicolumn{3}{c|}{RefCOCO}  & \multicolumn{2}{c}{RefCOCOg} \\
		20\% data &80\% data &val &testA &testB &val &test\\
		\hline
		Fully-anno &\ding{55}  &82.07 &87.33 &73.20 &68.08 &66.65  \\
		Fully-anno & Pseudo labels  &84.13 &88.36 &76.58 &74.64 &75.10  \\
		Fully-anno & Weakly (CyCo)  &86.73 &90.27 &81.87 &76.85 &77.81  \\
		Fully-anno & Fully-anno  &\bf{88.73} &\bf{91.07} &\bf{84.27} &\bf{80.23} &\bf{80.06}  \\
		\hline
	\end{tabular*}
	
	\vspace{+0.07in}
	
	\begin{tabular*}{8.4 cm} {@{\extracolsep{\fill}}cc|cc|cc}
		\hline
		\multicolumn{6}{c}{Ablation Study of Regional Image Captioning} \\
		\hline
		\multicolumn{2}{c|}{Training Data}  & \multicolumn{2}{c|}{RefCOCO testA}  & \multicolumn{2}{c}{RefCOCO testB} \\
		20\% data &80\% data &M &C &M &C\\
		\hline
		Fully-anno &\ding{55}  &29.6 &77.5  &34.0 &131.8  \\
		Fully-anno & Pseudo labels  &31.3 &87.4 &35.2 & 143.5  \\
		Fully-anno & Weakly (CyCo)   &33.7 &98.5  &36.6 &151.6  \\
		Fully-anno & Fully-anno  &{\bf 34.8} &{\bf 104.2} &{\bf 37.9} &{\bf 156.6}  \\
		\hline
	\end{tabular*}
\end{center}
\end{table}

\setlength{\tabcolsep}{2pt}
\begin{table}
\small
\begin{center}
	\caption{Performance study by adding cycle-consistency learning to the naive supervised VG model. Our ``supervised+cycle'' utilizes the \emph{same} training splits but adds cycle-consistency losses as the additional regularizations, which steadily improves the results.} \label{table:ablative on cycle}	
	\begin{tabular*}{8.4 cm} {@{\extracolsep{\fill}}cc|cc|cc|cc}
		\hline
		\multirow{2}{*}{Supervised}  &\multirow{2}{*}{Cycle}   & \multicolumn{2}{c|}{RefCOCO}  & \multicolumn{2}{c|}{RefCOCO+}  & \multicolumn{2}{c}{RefCOCOg} \\
		& &testA &testB &testA &testB &val &test\\
		\hline
		\checkmark &   &91.07 &84.27  &84.40 &68.20 &80.23 &80.06\\
		\checkmark  & \checkmark  &{\bf 91.87} &{\bf 85.33} &{\bf 87.07} &{\bf 69.87} &{\bf 81.31} &{\bf 81.04}\\
		\hline
	\end{tabular*}
\end{center}
\end{table}

%
%

{\noindent \bf Fully Supervised Training.}
%
%
We further combine this cycle-consistency learning pipeline with the standard supervised learning.
In each training iteration, we not only use the ground-truth labels to supervise the grounding model, but also utilize the \texttt{VG} $\rightarrow$ \texttt{IC} and \texttt{IC} $\rightarrow$ \texttt{VG} cycles to regularize the training.
As shown in Table~\ref{table:ablative on cycle}, based on the \emph{same} training data, adding cycle-consistency learning to the classic supervised learning can further boost performance.
This reveals that our framework can better exploit the training data by online data augmentation.

{\noindent \bf Adding More Training Data For VG.}
Since the proposed framework jointly optimizes visual grounding and image captioning within a unified framework, we can merge the datasets of both tasks for optimization.
As shown in Table~\ref{table:ablative on data combination}, adding captioning data further improves the visual grounding results.
Note that COCO Karpathy split lacks the region/box annotation, which can be regarded as the weakly-labeled data compared to the image-text-box triplet.
These results justify the potential of our framework of absorbing more cheap data to achieve superior performance.

{\noindent \bf Adding More Training Data For IC.}
In Table~\ref{table:ablative on image captioning}, we exhibit the image captioning potentials of our CyCo compared to the baseline BLIP.
Without the design of regional embedding, BLIP is not aware of the local region. 
To this end, combining both COCO-caption and RefCOCOg datasets fails to improve its performance.   
In contrast, our model can freely switch between the local and global captions (Figure~\ref{fig:ic_example}) and absorb more (fully- or partially-annotated) data to further boost the performance.

\setlength{\tabcolsep}{2pt}
\begin{table}
\small
\begin{center}
	\caption{Performance study by adding more training data. Adding weakly labeled images (\emph{e.g.,} COCO-caption without box annotations) improves the grounding performance.} \label{table:ablative on data combination}	
	\begin{tabular*}{8.3 cm} {@{\extracolsep{\fill}}cc|ccc|cc}
		\hline
		~\multirow{2}{*}{Ground Data}  &\multirow{2}{*}{Caption Data}   & \multicolumn{3}{c|}{RefCOCO}   & \multicolumn{2}{c}{RefCOCOg} \\
		& &val &testA &testB &val &test \\
		\hline
		\checkmark  &   &89.47 &91.87 &85.33 &81.31 &81.04 \\
		\checkmark  & \checkmark &{\bf 89.78} &{\bf 92.05} &{\bf 85.63} &{\bf 82.24} &{\bf 82.20} \\
		\hline
	\end{tabular*}
\end{center}
\end{table}

\setlength{\tabcolsep}{2pt}
\begin{table}
\small
\begin{center}
	\caption{Performance study on image captioning. Compared to our baseline, CyCo can better leverage the datasets from different domains (COCO-caption and RefCOCOg). } \label{table:ablative on image captioning}
	\begin{tabular*}{8.3 cm} {@{\extracolsep{\fill}}lccc}
		\hline
		&  Training Data  &RefCOCOg & COCO-caption \\
		&    &B@4~/~CIDEr &B@4~/~CIDEr \\
		\hline
		BLIP  & only COCO  & 24.1~/~74.6 &39.7~/~133.3 \\
		BLIP & COCO~\&~RefCOCOg  &30.8~/~96.2 &38.4~/~130.2\\
		{\bf CyCo} & COCO~\&~RefCOCOg   &{\bf 36.6~/~128.5} & {\bf 40.6~/~133.9} \\
		\hline
	\end{tabular*}
\end{center}
\end{table}

\begin{table*}[h]
\small
\begin{center}
	\caption{Comparisons with state-of-the-art methods on RefCOCO, RefCOCO+, and RefCOCOg in terms of top-1 accuracy (\%). The $\text{BLIP-B}^{\star}$ denotes our implemented BLIP-B model on the visual grounding task, which is not included in the original BLIP framework \cite{BLIP}. In the pre-training stage, we only use the image-text (I-T) pairs without any image-text-box (I-T-B) triplets. We report the performance of our semi-weakly supervised model as well as the supervised model.} \label{table:SOTA on RefCOCO}
	\begin{tabular*}{17.7 cm} {@{\extracolsep{\fill}}l|c|c|cc|ccc|ccc|cc}
		\hline
		~\multirow{2}{*}{Method}  &\multirow{2}{*}{Backbone}  &\multirow{2}{*}{Supervised}  &\multicolumn{2}{c|}{Data}   & \multicolumn{3}{c|}{RefCOCO}  & \multicolumn{3}{c|}{RefCOCO+} & \multicolumn{2}{c}{RefCOCOg} \\
		& &  &I-T &I-T-B &val &testA &testB &val &testA  &testB &val &test \\
		\hline
		~{\emph { w/o Pre-training}}  & & & & & & & & & & & &\\
		~MCN \cite{MCN}  &DarkNet-53 &\checkmark & & &80.08 &82.29 &74.98 &67.16 &72.86 &57.31 &66.46 &66.01\\
		~QRNet \cite{QRNet}   &Swin-S &\checkmark  & &  &84.01 &85.85 &82.34 &72.94 &76.17 &63.81 &73.03 &72.52 \\
		~VLTVG \cite{VLTVG}    &ResNet-101 &\checkmark  & &  &84.77 &87.24 &80.49 &74.19 &78.93 &65.17 &76.04 &74.18 \\
		~VG-LAW \cite{su2023language}   &ViT-B &\checkmark  & &  &86.62   &89.32 &83.16 &76.37 &81.04 &67.50 &76.90 &76.96 \\
        ~TransVG++ \cite{deng2023transvg++}  &ViT-B &\checkmark  & &  &86.28 &88.37 &80.97 &75.39 &80.45 &66.28 &76.18 &76.30\\
		\hline
		~{\emph {Img-Text Pre-training}} & & & & & & & & & & & &\\
        ~UNITER-B \cite{UNITER}   &ResNet-101 &\checkmark  &\checkmark &  &81.24 &86.48 &73.94 &75.31 &81.30 &65.58 &74.31 &74.51\\
		~VILLA-B \cite{VILLA}  &ResNet-101 &\checkmark  &\checkmark & &81.65 &87.40 &74.48 &76.05 &81.65 &65.70 &75.90 &75.93\\
		~ERNIE-ViL-B \cite{Ernie-vil}   &ResNet-101 &\checkmark  &\checkmark &  &- &- &- &74.02 &80.33 &67.74 &- &-\\
		~$\text{BLIP-B}^{\star}$ \cite{BLIP}   &ViT-B &\checkmark  &\checkmark &  &88.73 &91.07 &84.27 &79.53 &84.40 &68.20 &80.23 &80.06\\
		
		\hline
		~{\emph {Img-Text-Box Pre-training}} & & & & & & & & & & & &\\
		~\color{lightgray} MDETR \cite{MDETR}  &\color{lightgray} ResNet-101 &\color{lightgray} \checkmark &\color{lightgray} \checkmark &\color{lightgray} \checkmark  &\color{lightgray} 86.75 &\color{lightgray} 89.58 &\color{lightgray} 81.41 &\color{lightgray} 79.52 &\color{lightgray} 84.09 &\color{lightgray} 70.62 &\color{lightgray} 81.64 &\color{lightgray} 83.31\\
		
		~\color{lightgray} OFA-L \cite{OFA}   &\color{lightgray} BART-L  &\color{lightgray} \checkmark &\color{lightgray} \checkmark &\color{lightgray} \checkmark  &\color{lightgray} 90.05 &\color{lightgray} 92.93 &\color{lightgray} 85.26 &\color{lightgray} 84.49 &\color{lightgray} 90.10 &\color{lightgray} 77.77 &\color{lightgray} 84.54 &\color{lightgray} 85.20\\
		
		~\color{lightgray} FIBER-B \cite{FIBER}  &\color{lightgray} Swin-B & \color{lightgray} \checkmark &\color{lightgray} \checkmark &\color{lightgray} \checkmark &\color{lightgray} 90.68 &\color{lightgray} 92.59 &\color{lightgray} 87.26 &\color{lightgray} 85.74 &\color{lightgray} 90.13 &\color{lightgray} 79.38 &\color{lightgray} 87.11 &\color{lightgray} 87.32\\
		
		\hline
		
		~{\bf CyCo} (semi-weakly)  &ViT-B &\ding{55}  &\checkmark &  &86.73 &90.27 &81.87 &72.53 &82.00 &61.33 &76.85 &77.81 \\
		
		~{\bf CyCo} (fully supervised) &ViT-B &\checkmark &\checkmark &  &{\bf 89.47} &{\bf 91.87} &{\bf 85.33} &{\bf 80.40} &{\bf 87.07} &{\bf 69.87} &{\bf 81.31} &{\bf 81.04} \\
		
		\hline
	\end{tabular*}
\end{center}
\end{table*}

\subsection{State-of-the-art Comparison}

{\noindent \bf Evaluation on RefCOCO/RefCOCO+/RefCOCOg.}
Table~\ref{table:SOTA on RefCOCO} reports the comparison results of state-of-the-art methods on the RefCOCO, RefCOCO+, and RefCOCOg datasets.
As ablated in Table~\ref{table:ablative on data combination}, adding more captioning data can steadily boost the grounding performance.
For fairness, when fine-tuning the visual grounding model, we only use the standard training splits of the above benchmarks without involving other training data.
%
%
Thanks to the vision-language pre-training, our approach significantly outperforms the classic grounding methods without pre-training.
We also include some representative pre-training-based approaches including UNITER, VILLA, ERNIE-ViL, and BLIP to justify the superior performance of our approach.
For example, our method surpasses VILLA and ERNIE-ViL on the RefCOCO and RefCOCO+.
On the RefCOCOg dataset with longer expressions, our method also exhibits satisfactory results.
In Table~\ref{table:SOTA on RefCOCO}, we also present some high-performance methods (\emph{e.g.,} MDETR, OFA, and FIBER) that additionally leverage the expensive image-text-box data for pre-training.
Besides, FIBER adopts a strong visual backbone and higher image resolution (\emph{e.g.,} $800\times 1,333$), exhibiting the leading performance. 
We can also improve our performance by adopting a stronger backbone and higher image resolution, which leave as our future work.

Furthermore, our framework enables model training using partially labeled data. 
In the semi-weakly supervised setting (\emph{e.g.,} 20\% fully-annotated data and 80\% weakly-annotated data), it is worth noting that our method also steadily outperforms the previous fully supervised counterparts such as UNITER and VILLA on the RefCOCO and RefCOCOg.
%
%

%
%
%
%

\setlength{\tabcolsep}{2pt}
\begin{table}[t]
\small
\begin{center}
	\caption{Performance comparisons on the COCO-caption Karpathy test split, where B@4, M, C, S denote BLEU@4, METEOR, CIDEr, and SPICE scores, respectively.} \label{table:SOTA on COCO Caption}
	\begin{tabular*}{8.2 cm} {@{\extracolsep{\fill}}l|cccc}
		\hline
		~\multirow{2}{*}{Method}   & \multicolumn{4}{c}{Cross-Entropy}  \\
		&B@4 &M &C &S \\
		\hline
		~{\emph{w/o Pre-training}} & &  &  &  \\
		~AoANet \cite{AoANet}    &37.2 &28.4 &119.8 &21.3  \\
		~X-LAN \cite{XLAN}   &38.2 &28.8 &122.0 &21.9  \\
		\hline
		~{\emph{w/ Pre-training} } &&  &  &  \\
		~Oscar-B \cite{OSCAR}    &36.5 &30.3 &123.7 &23.1  \\
		~VinVL-B \cite{VINVL}  &38.2 &30.3 &129.3 &23.6  \\
		~LEMON-B \cite{LEMON}  &40.3 &30.2 &133.3 &23.3  \\
		~BLIP-B \cite{BLIP}    &39.7 &- &133.3 &23.3  \\
		~SimVLM-B \cite{SimVLM}   &39.0 &{\bf 32.9} &{\bf 134.8}  &24.0 \\
		~{\bf CyCo (Ours)}  &{\bf 40.6}  &31.2  &133.9  &{\bf 24.4}\\
		\hline
	\end{tabular*}
\end{center}
\end{table}


{\noindent \bf Evaluation on COCO Caption.}
Mainstream image captioning methods focus on describing the global image context.
However, optimizing the proposed framework only on the grounding dataset will overlook the captioning capability of the global content.
To this end, we merge two datasets including COCO Karpathy train split and RefCOCOg train split to jointly optimize our cycle-consistency framework.
We define the referred region of COCO-caption dataset as the whole image.
After joint training, we assess the global captioning performance on the COCO Karpathy test split.
%

%
%

In Table~\ref{table:SOTA on COCO Caption}, compared with the state-of-the-art captioning methods with pre-training such as Oscar and VinVL, our method shows better results.
Our method is even comparable with the recent SimVLM trained with 1.8 billion image-text pairs, which is 10$\times$ larger than ours. 
The proposed method slightly outperforms our baseline BLIP on the COCO (global image) captioning task.

\begin{figure}
\centering
\includegraphics[width=8.4cm]{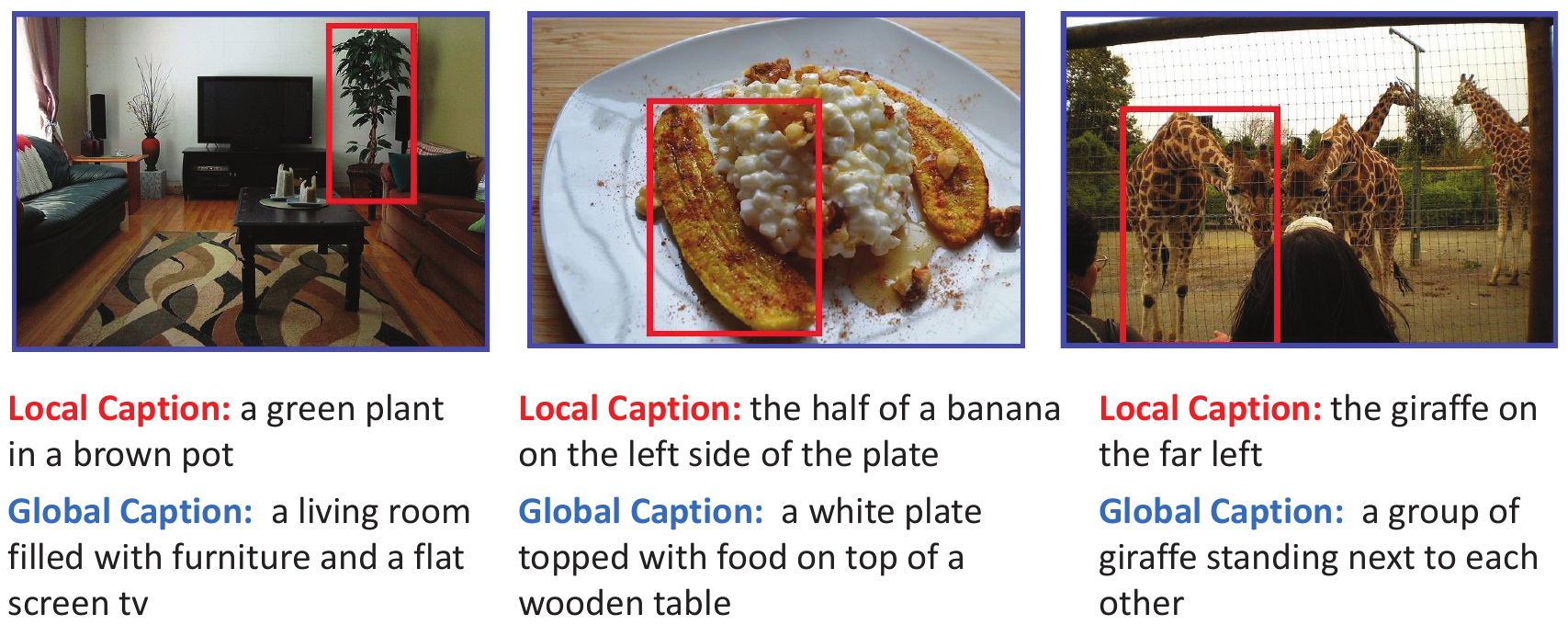}
\caption{Our captioner can describe both the local ({\color{red} red region}) and global ({\color{blue} blue region}) contexts.}
\label{fig:ic_example}
\end{figure}


%
%
%

%
%
%

It is worth noting that our captioner focuses on a more challenging scenario and can describe arbitrary image regions, which is infeasible for all the comparison methods.
Figure~\ref{fig:ic_example} showcases some examples.
When describing a specific region, our captioner captures its relationship with surrounding objects to avoid ambiguous expressions.
When describing the global image content, the proposed method performs in a similar way as the general image captioners.

\section{Conclusion}

In this paper, we bridge two vision-language tasks including visual grounding and regional image captioning in a cyclic training pipeline.
By cycle-consistency constraints, the proposed framework can exploit the weakly-annotated data for model training and augment the training samples to further boost the fully supervised training.
Extensive experiments on image captioning and visual grounding datasets verify the effectiveness of our framework.

{	
\small
\bibliographystyle{aaai}
\bibliography{aaai24}
}

\end{document}